\documentclass[letterpaper, 10 pt, conference]{ieeeconf}
\IEEEoverridecommandlockouts
\usepackage{cite}
\usepackage{amsmath,amssymb,amsfonts}
\usepackage{algorithmic}
\usepackage{graphicx}
\usepackage{textcomp}
\usepackage{xcolor}
\usepackage{caption}
\usepackage{booktabs}
\usepackage{makecell}
\usepackage{hyperref}

\usepackage{subcaption}
\graphicspath{ {./figures/} }

\newcommand{\tb}[1]{\textcolor{black}{#1}}

\def\BibTeX{{\rm B\kern-.05em{\sc i\kern-.025em b}\kern-.08em
    T\kern-.1667em\lower.7ex\hbox{E}\kern-.125emX}}
\begin{document}


\title{Multi S-Graphs: An Efficient Distributed \\Semantic-Relational Collaborative SLAM}


\author{
    Miguel Fernandez-Cortizas$^{1,2}$, Hriday Bavle$^{1}$, David Perez-Saura$^{2}$,\\ \;\;Jose Luis Sanchez-Lopez$^{1}$, Pascual Campoy$^{2}$ and Holger Voos$^{1}$ 
    \thanks{$^{1}$Authors are with the Automation and Robotics Research Group, Interdisciplinary Centre for Security, Reliability, and Trust (SnT), University of Luxembourg, Luxembourg. Holger Voos is also associated with the Faculty of Science, Technology, and Medicine, University of Luxembourg, Luxembourg. \tt{\small{\{hriday.bavle, joseluis.sanchezlopez, holger.voos\}}@uni.lu}}
    \thanks{$^{2}$Authors are with the Computer Vision and Aerial Robotics Group at Universidad Politécnica de Madrid, Spain (CVAR-UPM) at the Centre for Automation and Robotics C.A.R. (UPM-CSIC). \tt\small{\{miguel.fernandez.cortizas, david.perez.saura, pascual.campoy\}@upm.es}}
    \thanks{*This work was partially funded by the Fonds National de la Recherche of Luxembourg (FNR), under the projects C22/IS/17387634/DEUS, by European Union's Horizon Europe Project No. 101070254 CORESENSE, as well as project COPILOT ref. 2020/EMT6368, funded by the Madrid Government under the R\&D Synergic Projects Program, project INSERTION ref. ID2021-127648OBC32 and project RATEC ref: PDC2022-133643-C22, both funded by the Spanish Ministry of Science and Innovation. The work of the third author is supported by the Spanish Ministry of Science and Innovation under its Program for Technical Assistants PTA2021-020671.}
    \thanks{For the purpose of Open Access, and in fulfillment of the obligations arising from the grant agreement, the author has applied a Creative Commons Attribution 4.0 International (CC BY 4.0) license to any Author Accepted Manuscript version arising from this submission.}
}
\maketitle
\begin{abstract}

Collaborative Simultaneous Localization and Mapping (CSLAM) is critical to enable multiple robots to operate in complex environments. Most CSLAM techniques rely on raw sensor measurement or low-level features such as keyframe descriptors, which can lead to wrong loop closures due to the lack of deep understanding of the environment. Moreover, the exchange of these measurements and low-level features among the robots requires the transmission of a significant amount of data, which limits the scalability of the system. To overcome these limitations, we present \mbox{\textit{Multi S-Graphs}}, a decentralized CSLAM system that utilizes high-level semantic-relational information embedded in the four-layered hierarchical and optimizable situational graphs for cooperative map generation and localization \tb{in structured environments} while minimizing the information exchanged between the robots. To support this, we present a novel room-based descriptor which, along with its connected walls, is used to perform inter-robot loop closures, addressing the challenges of multi-robot kidnapped problem initialization. Multiple experiments in simulated and real environments validate the improvement in accuracy and robustness of the proposed approach while reducing the amount of data exchanged between robots compared to other state-of-the-art approaches.
\\
Software as docker image: \url{https://github.com/snt-arg/multi_s_graphs_docker}
\\
\tb{Video: \url{https://youtu.be/0JsDn651rdc}
}

\end{abstract}

\section{Introduction}

\begin{figure}
\centering
    \includegraphics[width=0.4\textwidth]{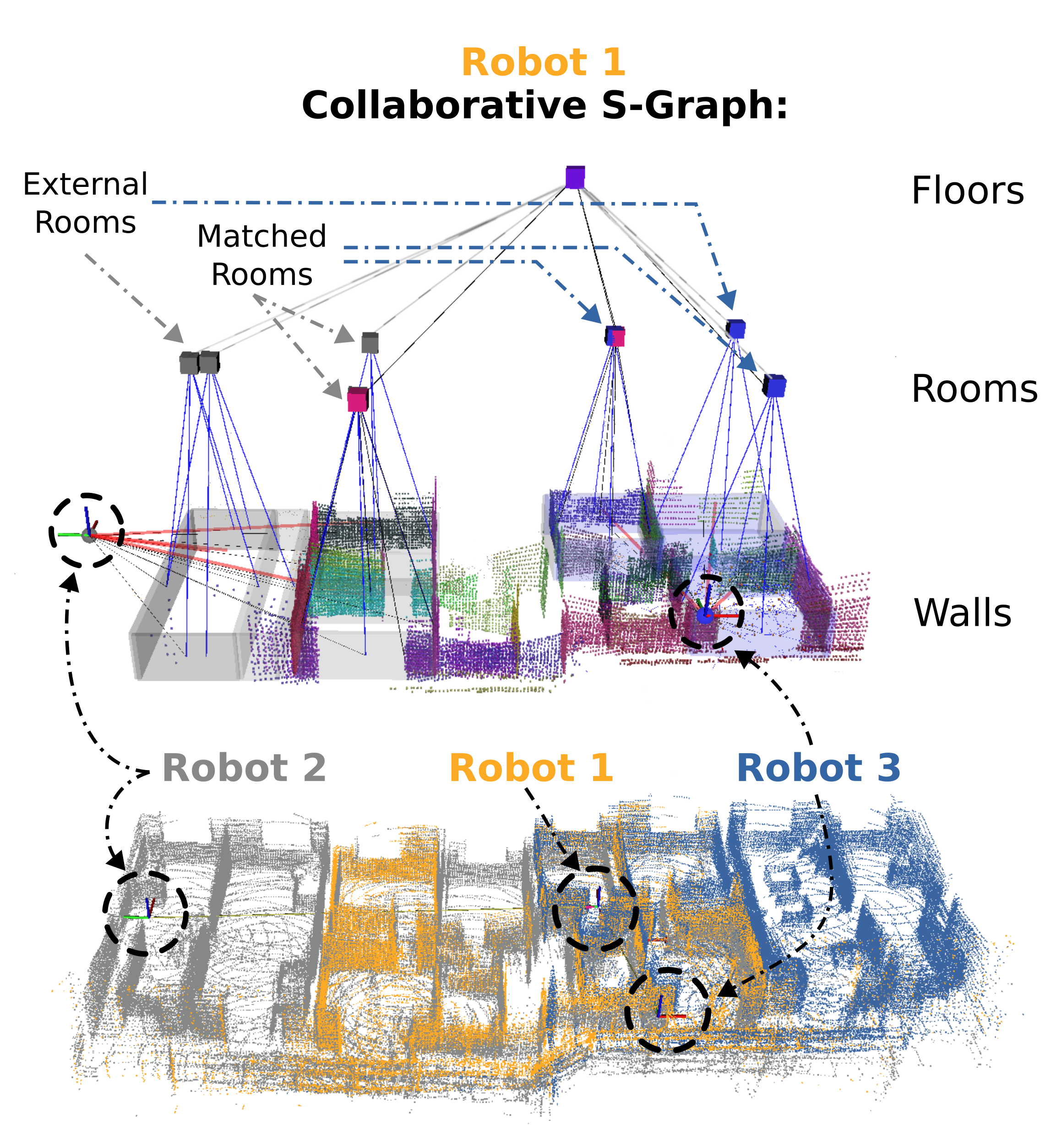}
   \caption{Collaborative graph generated by three robots mapping a building floor (orange, gray, and blue for robots 1, 2, and 3 respectively). The collaborative graph is generated from the point of view of Robot 1. Semantic information gathered by the other robots, such as rooms or walls appears in transparent colors in the graph layers. The origin of the coordinate systems of the other robots appears surrounded by a circle. Below is the result of merging the point clouds obtained from each of the robots. This point cloud is displayed for visualization purposes and is not utilized in inter-robot communication.}
    \label{fig:figure1}
    \vspace{-0.4cm}
\end{figure}

Collaborative Simultaneous Localization and Mapping (CSLAM) is a technology that enables multi-robot systems to work together to create a map of their environment while simultaneously determining their own positions within the map.
Most CSLAM techniques, such as \cite{Lajoie2020},\cite{zhong2022dclslam}, \cite{Huang2022disco}, are highly based on the transmission of low-level features, such as keyframe descriptors, between robots. With these low-level features, each robot creates its own map and can match it to integrate them into a common map. This is used for both visual and LiDAR-based approaches.
However, the use of these low-level features usually leads to incorrect loop closures \cite{azzam2020feature}. Works such as \cite{Lajoie2020}, \cite{mangelson2018pairwise} focus on robustifying their loop closure algorithms to avoid incorrect loop closures. The main drawback of these methods emerges from the fact that the system has limited understanding of the environment.



Novel approaches to single-robot SLAM address this issue of limited understanding of the environment by incorporating meaningful metric-semantic-relational information, with works like {Hydra} \cite{Hughes2022} which exploits 3D scene graphs for loop closure, or \textit{S-Graphs+} \cite{bavle2022a} \cite{Bavle2022}, which couple in a situational graph modeled as a single optimizable factor graph, the 3D scene graph with the underlying SLAM graph.
These works demonstrate that extracting semantic meaning from metric information and incorporating it with appropriate relational constraints such as walls, rooms, floors, etc., improve the overall precision of underlying SLAM while robustifying loop closures and generating meaningful environmental maps. 

Recently, Hydra-Multi \cite{chang2023hydramulti} and CURB-SG \cite{greve2023collaborative} took advantage of 3D scene graphs for collaborative SLAM in a centralized way. These approaches harness semantic-relational knowledge to exchange higher-level information between the agents to fortify inter-robot loop closures.  However, these methods, apart from being centralized, explicitly add loop closure constraints to underlying robot poses instead of implicitly constraining the same semantic entities detected by different agents.

Thus in this work, we present \textit{Multi S-Graphs}, a LiDAR-based distributed CSLAM system \tb{for structured environments} harnessing the potential of the multi-layered Situational Graphs (S-Graphs), a single optimizable factor graph that combines a pose graph (i.e. with keyframes) and a 3D scene graph with semantic-relational concepts, to generate precise environmental maps and locate the different robots in this map while exchanging minimum information between robots.
To the best of our knowledge, we present the first distributed collaborative SLAM system based on optimizable situational graphs. Fig. \ref{fig:figure1} shows the \textit{Multi S-Graphs} executed on a real experiment. 

The main contributions of our system are as follows:
\begin{enumerate}
    \item A novel distributed multi-robot framework, \textit{Multi S-Graphs} based on four-layered optimizable hierarchical S-Graphs, exploiting semantic-relational information for map merging while minimizing the information exchange between the robots.
    \item \tb{A hybrid room-based descriptor combining raw point cloud information with semantic and hierarchical knowledge to identify inter-robot loop closures.}  
    \item Experimental validation of the proposed approach in simulated and real environments obtaining improved accuracy compared to state-of-the-art multi-robot approaches.
\end{enumerate}

\section{Related Work}



\subsection{Multi-Agent SLAM}


\textbf{Centralized.}
Centralized CSLAM approaches are based on multiple single-robot front-ends with a common back-end.
LAMP 2.0 \cite{chang2022lamp} is a centralized multi-robot SLAM system that is adaptable to different input odometry sources, with a robust and scalable loop closure detection module and an outlier-robust back-end based on Graduated Non-Convexity (GNC) for pose graph optimization. It is designed for operation in challenging large-scale underground environments. Maplab 2.0 \cite{maplab2023} is a large-scale multi-robot multi-session mapping system that integrates non-visual landmarks and incorporates a semantic object-based loop closure module into the mapping framework. COVINS-G \cite{covinsg2023} is a generalized back-end for collaborative visual-inertial SLAM building upon the COVINS \cite{covins2021} framework that enables compatibility with any arbitrary VIO front-end. The drawbacks of centralized architectures are that they require good communication with the common back-end and, if that node fails, the whole system fails. In addition, they do not scale up for large robot teams.

\textbf{Decentralized.}
Within decentralized approaches, most of them focus on visual-based SLAM. Deutsch et al. \cite{Deutsch2016} proposed a framework that relies on a Bag of Words (BoW) of the keyframes obtained with an RGB-D camera to enable collaborative mapping for existing (single-robot) SLAM systems. 
Kimera-Multi \cite{kimeramulti2022} also uses BoW and needs a \textit{Robust Distributed Initialization} to initialize all robot poses in a shared (global) coordinate frame. It builds a globally consistent metric-semantic 3D mesh model of the environment in real time, while being robust to incorrect loop closures and only relying on local communication. Each robot builds a local trajectory estimate and a local mesh using Kimera \cite{kimera2020}, and when two robots are within communication range, they initiate a distributed place recognition and robust pose graph optimization protocol based on a distributed GNC algorithm. 
Swarm-SLAM \cite{swarmslam2023} includes a novel inter-robot loop closure prioritization technique based on algebraic connectivity maximization that reduces communication and accelerates convergence. It also has a decentralized approach to neighbor management and pose graph optimization suited for sporadic inter-robot communication, and supports different types of sensors using different descriptors, such as CosPlace \cite{cosplace2022} and NetVLAD \cite{netvlad2018} for images and Scan Context \cite{kim2018scan} for LiDAR scans.
Decentralized LiDAR SLAM solutions focus on the data-efficient exchange of observation between robots. 
DiSCo-SLAM \cite{Huang2022disco} uses the lightweight Scan Context descriptor and includes a two-stage global and local optimization framework for distributed multi-robot SLAM which provides stable localization results that are resilient to the unknown initial conditions that typify the search for inter-robot loop closures. 
DCL-SLAM \cite{zhong2022dclslam} includes a lightweight global descriptor, LiDAR-Iris \cite{lidariris2020}, for loop closure detection, a three-stage data-efficient distributed loop closure approach to estimate the relative pose transformation between robots, and a two-stage distributed Gauss-Seidel (DGS) approach for optimization.
\begin{figure*}[!t]
    \centering
    \vspace{-0.2cm}
    \includegraphics[width=0.79\textwidth]{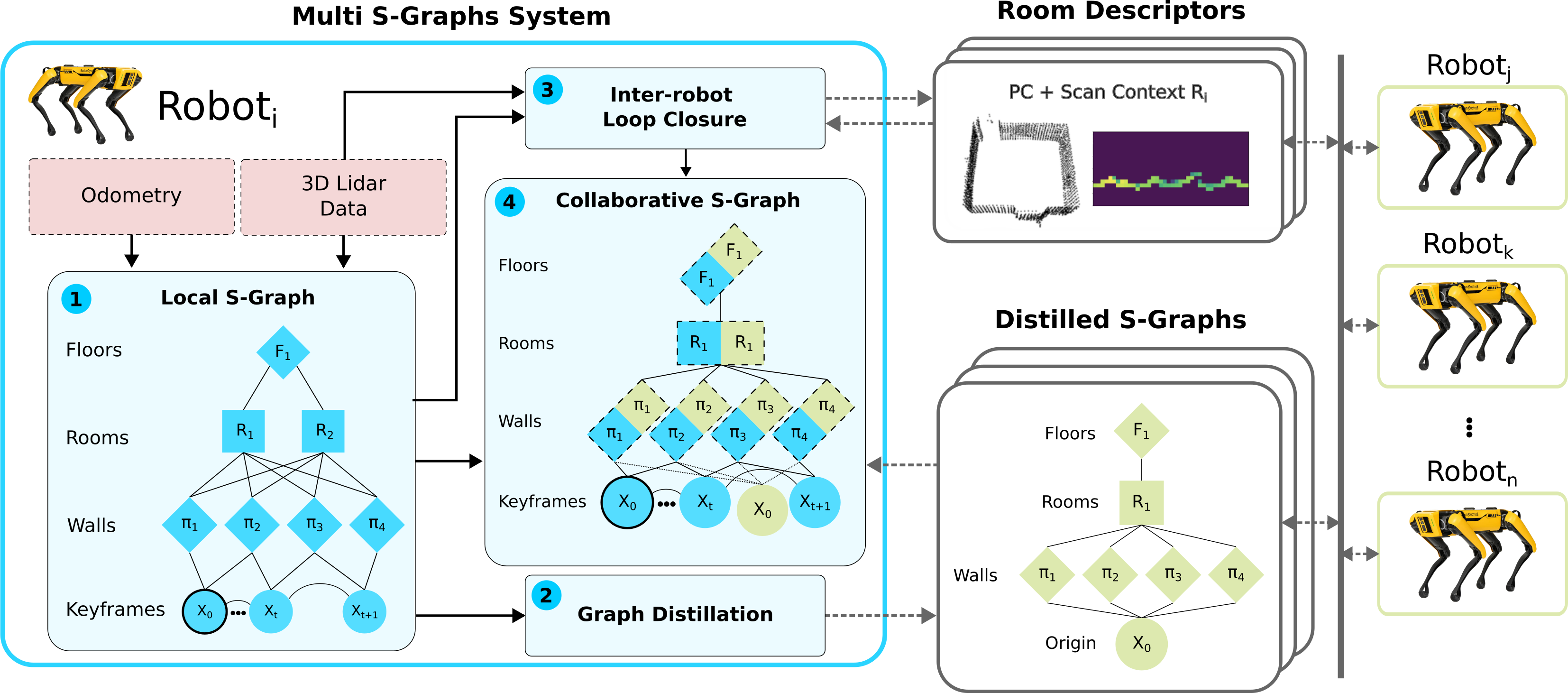}
    \caption{\textit{Multi S-Graph} architecture viewed from $Robot_{i}$ perspective. Odometry measures and 3D Lidar data are the inputs that the robotic platform provides. Each robot executes its own Local S-Graph system and generates the Room Descriptors and the Distilled S-Graphs that are exchanged with the rest of the \tb{robots}. Messages exchanged between robots are represented in gray. From the point of view of the $Robot_{i}$ the rest of the agents provide the Distilled graphs and the Room descriptors that are used to generate the Collaborative S-Graph. Information provided by the rest of the robots appears in green color.}
    \label{fig:multi_sgraph_schema}
    \vspace{-0.6cm}
\end{figure*}
Current decentralized collaborative SLAM systems struggle to effectively use well-mapped areas generated by other agents, leading to map redundancy and costly re-mapping. Bänninger et al \cite{cross-agent2023} propose a strategy to efficiently share areas mapped by different agents in a decentralized, peer-to-peer fashion, achieving a globally consistent estimate through distributed bundle adjustment using the Alternating Direction Method of Multipliers (ADMM). The method leverages covisibility information between keyframes instantiated by different agents to transfer local sub-maps on-the-fly, addressing map redundancy while maintaining consistency and scalability. However, the use of semantic information from the environment can further reduce the information exchange.

\subsection{Semantic Information in CSLAM}


\tb{One of the challenges in CSLAM is combining the local maps of each robot into one global one when the initial 
pose of each one is unknown. Single-robot SLAM approaches like \cite{yu2022semanticloop},\cite{tao2024active},\cite{segmap2018}, utilize semantic object detection, enhancing the robustness and accuracy by reducing the loop closure search space. Motivated by this, some CSLAM  approaches like \cite{yu2020review},\cite{vascak2023mapmerging} utilize object-level semantics to robustify inter-robot map merging. In terms of map fusion, \cite{frey2019constellation} presented a framework for efficient map-merging in semantic object-based SLAM detecting duplicate constellations of landmarks. In \cite{prabhu2024forestry}, the authors utilized semantic place recognition on sparse representations of tree trunks, represented by cylinder models extracted from a LIDAR sensor, to calculate relative transformations between multiple robots. All these works do not utilize hierarchical information along with semantic information to enhance inter-robot map merging as well as require densely populated indoor scenes with relevant objects.}

\subsection{Multi-Agent 3D Scene Graphs}
Recent research aims to leverage the use of semantic information by combining Scene Graphs with SLAM. Hydra-Multi \cite{chang2023hydramulti} is the first multi-robot spatial perception system capable of building a hierarchical 3D scene graph online from different sensor data. Based on Hydra \cite{Hughes2022}, it is a centralized architecture that includes a hierarchical loop closure detection module based on 3D scene graphs, a frame alignment module that estimates the initial poses of the robots, and an alignment-optimize-reconcile framework that uses GNC to optimize multi-robot 3D scene graphs.
Recently, CURB-SG \cite{greve2023collaborative} has been presented as the first solution to create large-scale 3D scene graphs of dynamic urban scenes for automated driving, using collaborative mapping techniques and leveraging panoptic LiDAR data.
However, none of these approaches are decentralized, which limits their scalability.
\section{System Overview}




In \textit{Multi S-Graphs}, each robot runs the fully decentralized \textit{Multi S-Graphs} system as is shown in Fig \ref{fig:multi_sgraph_schema}. Our system architecture is built around a Situational Graph (S-Graph), a single multi-layered optimizable factor graph that combines a pose graph and a scene graph with semantic-relational concepts. This graph contains all the knowledge that is used for extracting the key information needed for exchange and integrates information from other agents, leading into a Multi-robot Situational Graph. The information that is shared between agents is broadcasted into a common communication channel that exchanges a minimal amount of data between them. In our system, we consider the multi-robot kidnapped problem, where each robot is unaware of its location on the map as well as the location of the other robots, 
so each robot creates a map of the environment with respect to its starting point (origin) and localizes the rest of the agents with respect to it.
Our proposed system is composed of four main modules:

\textbf{1. Local S-Graph Generation (Section \ref{sec::LS-Graph}).} To generate the local S-Graph for each robot, we use \tb{\textit{S-Graphs+}} \cite{Bavle2022} which runs onboard each one using the odometry and 3D LiDAR measurements of the robot. It also estimates the position and trajectory followed by the robot while generating it. \tb{Like \textit{S-Graphs+}, our system can work with any external odometry source.}

\textbf{2. S-Graph Distillation (Section \ref{sec::situational_graph_distillation}).}
We marginalize the local S-graph generated by each robot to only transmit semantic-relational information and not the low-level information stored on each keyframe. Each distilled graph includes a reference node that represents the origin of the local coordinated system of each robot. This distilled S-graph is transmitted to the rest of the robots.

\textbf{3. Room-Based Inter-Robot Loop Closure (Section \ref{sec:room_descriptors}).}
Relying only on the generated semantic graph for loop closure can generate alignment errors in the presence of symmetry in the environment. In order to combine the raw information of the point clouds and take advantage of the high-level semantic information that each room contains, we generate a hybrid descriptor for each room, a \textit{Room Descriptor}.  This module is in charge of two tasks: generating the descriptors and finding appropriate room matches.

\textbf{4. Collaborative S-Graph Generation and Optimization (Section \ref{sec:collaborative_s_graphs}).} This module integrates the Distilled S-Graphs received from the other agents with its own local S-Graph. The different graphs are connected together with the loop closures generated by the previous module. The connected graphs are globally optimized to generate the final collaborative situational graph.  






\section{Local Situational Graph}
\label{sec::LS-Graph}
In our system, the situational awareness of each robot is encapsulated within a situational graph whose entities have been designed to represent the different elements that make up the layout of a building. 
In this work, we rely on the \textit{S-Graphs+}\cite{Bavle2022} algorithm for generating this Situational Graph, based on the LiDAR and odometry measurements of the robots and estimating the localization of the robot. These S-Graphs are four-layered optimizable hierarchical graphs that contain the relevant information regarding the structure of the building:

\textbf{Keyframes Layer.} It consists of robot poses factored as SE(3) nodes in the agent map frame $A_i$ with pairwise odometry measurements constraining them.

\textbf{Walls Layer.} It consists of the planar wall surfaces extracted from the 3D LiDAR measurements and factored
using minimal plane parameterization. The planes observed by their respective keyframes are factored using pose-plane constraints.

\textbf{Rooms Layer.} It consists of two-wall rooms (corridors) or four-wall rooms, each constraining either two or four detected walls, respectively.

\textbf{Floors Layer.} It consists of a floor node positioned in the center of the current floor.

\tb{We will use the four-wall rooms for the S-Graph distillation (Section \ref{sec::situational_graph_distillation}) and the Room-based loop closure (Section \ref{sec:room_descriptors}) because the boundaries and dimensions of these rooms can be easily extracted from the walls that constitute them.}

\section{Situational Graph Distillation}

\label{sec::situational_graph_distillation}
To reduce the amount of data that has to be interchanged between robots, we extract the most meaningful information from the local S-Graph. This graph encodes higher-level layers of the S-Graph and locates them with respect to the origin node of each robot, which enables a minimal but meaningful information exchange about the structure of the building. 
The Distilled S-Graph is organized in the following layers:

    \textbf{Origin Layer.} The origin node of each Distilled S-Graph, it represents the origin of the coordinated system of each robot. All the rest of the elements of the graph are located with respect to this node.
    
    \textbf{Walls Layer.} The estimation and covariances of the walls extracted from the S-Graph of each agent. The edges connecting the planes with their corresponding keyframe nodes are replaced by an edge to the origin vertex. 
    
    \textbf{Rooms Layer.} The estimation of the four-wall room vertex detected by each agent and the edges connecting the rooms to their four respective wall nodes.
    
    \textbf{Floors Layer.} It consists of a floor node positioned in the center of the current floor.

\section{Room-Based Loop Closure}
\label{sec:room_descriptors}
In order to connect the Local S-Graph with the external Distilled S-Graphs a loop closure method is required.  To avoid errors aligning the robot positions in very symmetric situations, like a corridor with multiple rooms, one on side of the order, we cannot only rely on the structural information stored in the Distilled S-Graphs, more fine-grained information may be needed to break the symmetry and decide if two rooms are the same or not.
\subsection{Room descriptor generator}
\begin{figure}[!h]
    \centering
    \includegraphics[width=0.48\textwidth]{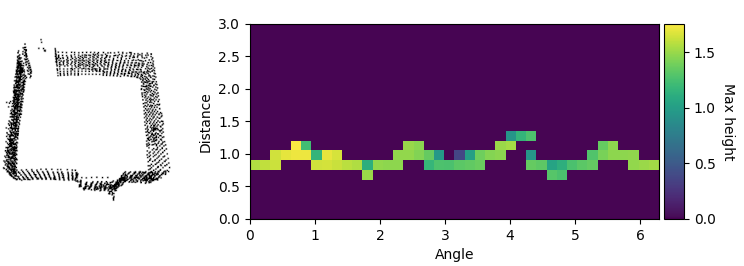}
    \caption{Room Descriptor (right) obtained from a Room Centric point cloud (left) by using a Scan Context. The value of each cell in the descriptor is the maximum height of the points within a corresponding bin.}
    \label{fig:room_descriptor}
\end{figure}

We present a hybrid descriptor, a Room Descriptor that integrates the semantic information of each room with a \textit{Room Centric point cloud} based descriptor, which enhances the robustness and accuracy of the loop closure algorithm by a small increase in the data transferred. An example of a Room Descriptor is shown in Fig. \ref{fig:room_descriptor}.

To generate these descriptors, we rely on the Scan Context descriptor \cite{kim2018scan}. This is an egocentric, yaw-invariant, point cloud descriptor that has demonstrated state-of-the-art results, given its simplicity and fast convergence time. 
The main drawback of this descriptor is that it is sensitive to linear motions.
\tb{To address this issue, we propose to generate Scan Contexts based on a Room Centric point cloud, which reduces the sensitivity to linear motion and strengthens descriptor matching.}

\tb{In this work, we exploit the semantic and hierarchical information from Local S-Graph to obtain the room center and its boundaries to extract all the keyframes obtained from within the room and generate its corresponding Room Centric point cloud.}



Each room keyframe $R_{k_i}$ can be expressed as:
\begin{equation}
    R_{k_{i}} = U \{^{R_{i}}K_j\} \quad ;\quad  \forall j \;|\; K_j \in R_{i}
\end{equation}
where $^{R_{i}}K_j$ represents the point cloud associated with the keyframe $K_j$ in the $R_i$ frame (a frame located in the center of the room $i$).

To obtain the room descriptor $R_{d_i}$ from a room keyframe $R_{k_i}$ we first down-sample $R_{k_i}$ with a fixed voxel size to homogenize the number of points for each room. 
Finally, the Scan Context descriptor of each room keyframe $R_{k_i}$ is computed to create the room descriptor $R_{d_i}$:
\begin{equation}
    R_{d_{i}} = SC(\phi(R_{k_i}))
\end{equation}
where $\phi(R_{k_i})$ represents the down-sampled room keyframe cloud, and $SC(\vec{X}):R^{n\times3} \rightarrow R^{n_s\times n_r}$ is the Scan Context function obtaining the descriptor for a given point cloud.

\subsection{Loop Closure detection}
\label{section:loop_closure}
In order to generate good candidates for alignment, we leverage the room descriptors to match rooms that are used for connecting the local S-Graph with the external distilled graphs, generating a global alignment of each robot coordinate system.
Room matching is performed in a two-step process:

\textbf{\textit{Scan Context Matching}:} Each robot receives and stores the room descriptors of the rest of the agents, trying to find a suitable match using the Scan Context distance \cite{kim2018scan}.

\textbf{\textit{Fine Alignment:}} Whenever a match is found between the robot and other agents' room, \tb{it tries to obtain an improved transform by asking for the Room Centric point cloud that originates the Scan Context and using a VGICP\cite{vgicp} registration algorithm.}

If a match is found, the room match with the relative transformation between rooms is sent into the following module (Section \ref{sec:collaborative_s_graphs}) to connect the different graphs. 



\tb{Scan Context and ICP match validity depend on the $SC_{th}$ and $ICP_{th}$ thresholds respectively.} \tb{The first one regulates the volume of point clouds sent over the network, while the second adjusts the match precision.} \tb{It is crucial for Room Descriptors to remain unique and constant over time for accurate matches, as identical rooms or changing environments can lead to erroneous matches.}


\section{Collaborative Situational Graph}
\label{sec:collaborative_s_graphs}

The final module integrates the Local S-Graph with the Distilled S-Graphs, using the loop closures provided by the previous module, generating a Collaborative S-Graph. During this Collaborative S-Graph generation, two different stages can be found: 

\textbf{Isolated Graphs}: At startup, this module stores the information from the Local S-Graph and the Distilled S-Graphs received by the other robots but, since the robots do not know where each one has started, it cannot combine the different graphs. At this stage, each external distilled S-Graph has its origin layer connected to the walls and rooms layer, but this origin layer is ``free", as there are no connections with the local S-Graph of the robot, resulting in multiple disconnected graphs. As long as new information is received, these graphs are updated.

\textbf{Collaborative Graph}: If a room match is found using the room matching step (Section \ref{sec:room_descriptors}), we formulate the factors between the local s-graph and the distilled s-graphs in the 2 steps. 
In the first step, a room-to-room factor is created for the matched rooms. For the given matched rooms, lower-level wall-to-wall factors are established, which robustify the alignment of the graphs. \tb{As seen in Fig \ref{fig:multi_sgraph_schema}, graphs are merged at rooms, walls, and floor levels. However, since floor nodes connect all the rooms at the same height, no explicit floor-to-floor factors are required between the floor nodes of the robots.}
Whenever local S-Graphs and external distilled S-Graphs are connected with appropriate factors, a common optimization is performed similarly to [8], \tb{leading into an optimized Collaborative S-Graph and, consequently, an optimized state estimation of the robot.}
After this optimization, the external graphs are aligned with the robot's local coordinate frame. \tb{The new information provided by the other robots can be integrated into the collaborative graph}, using the same coordinated frame as the local robot for representing the different entities of the external distilled graphs, as can be seen in Fig. \ref{fig:figure1}. \tb{Subsequently, if a new room match is found, the same two-step room factors are formulated, increasing the number of factors between graphs and refining the optimized graph.}

\tb{The Local S-Graph generated by \textit{S-Graphs+} is being updated continuously, adding new nodes into the graph and including intra-robot loop closures at the four hierarchical levels as in \cite{bavle2022a}. As the Collaborative Graph includes this local graph, these updates are also integrated into it.}
\section{Experimental Validation}

\subsection{Methodology}

We have selected the following state-of-the-art lidar-based distributed CSLAM algorithms for comparison due to the availability of their code: DCL-SLAM \cite{zhong2022dclslam}, Swarm-SLAM \cite{swarmslam2023}, and Disco-SLAM \cite{Huang2022disco}.
Additionally, we also challenge our algorithm against a centralized CSLAM, LAMP 2.0 \cite{chang2022lamp}, because centralized algorithms perform better than distributed ones when the number of robots is small.

The experiments were conducted with quadruped robots equipped with LiDAR in structured environments in the context of the ``multi-robot kidnapped problem", where the initial positions of the robots were unknown, with the exception of LAMP 2.0, as it requires prior knowledge of the initial positions of the robots within the environment. 

We conducted experiments in simulated and real-world scenarios. In simulation, we present our results in terms of the Absolute Trajectory Error (ATE) compared to the provided ground truth. In real-world experiments, we report the Root Mean Square Error (RMSE) of the estimated 3D maps compared to the actual 3D map derived from the architectural plans. We also performed an analysis of the data exchanged between the agents for each scenario. Along with this, we performed an ablation study to evaluate the contribution of the different hierarchical factors between the local S-Graph and the distilled S-Graphs in terms of accuracy and data transmission. 



\textbf{Simulated Data.}
We employed the Gazebo3 physics simulator to replicate indoor settings for our experiments. In all these simulated scenarios, we relied on LiDAR data for odometry estimation. Our study encompassed a total of three simulated experiments, out of which C1F0 and C1F2 were constructed using 3D mesh representations of real architectural plans, and SE1 is a simulated environment mimicking common indoor layouts with varying room configurations. 

\begin{figure}[!h]
\vspace{-0.2cm}
     \centering
     \begin{subfigure}[b]{0.15\textwidth}
        \centering
        \includegraphics[width=0.98\textwidth]{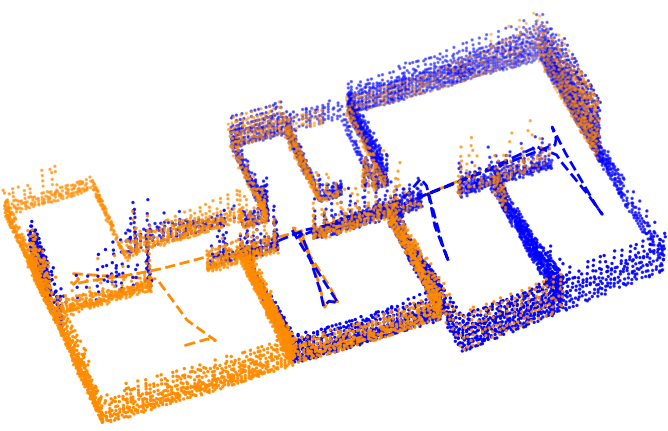}\\
        \caption{C1F0s}
        \label{fig:exp:sim:c1f0}
     \end{subfigure}
     \begin{subfigure}[b]{0.15\textwidth}
        \centering
        \includegraphics[width=0.98\textwidth]{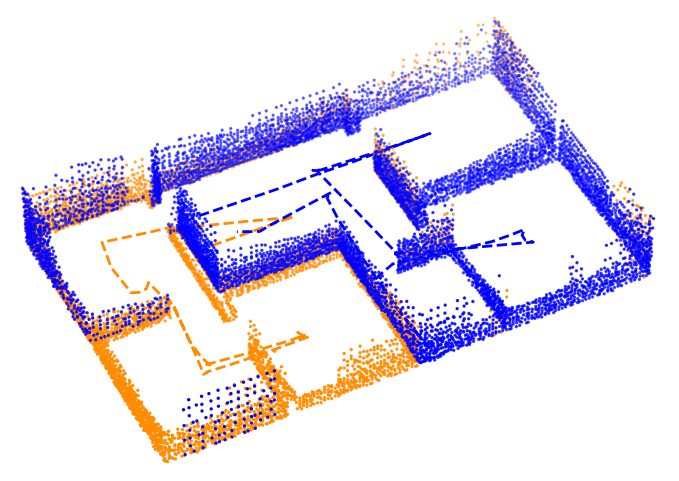}\\
        \caption{C1F2s}
        \label{fig:exp:sim:c1f1}
     \end{subfigure}
     \begin{subfigure}[b]{0.15\textwidth}
        \centering
        \includegraphics[width=0.98\textwidth]{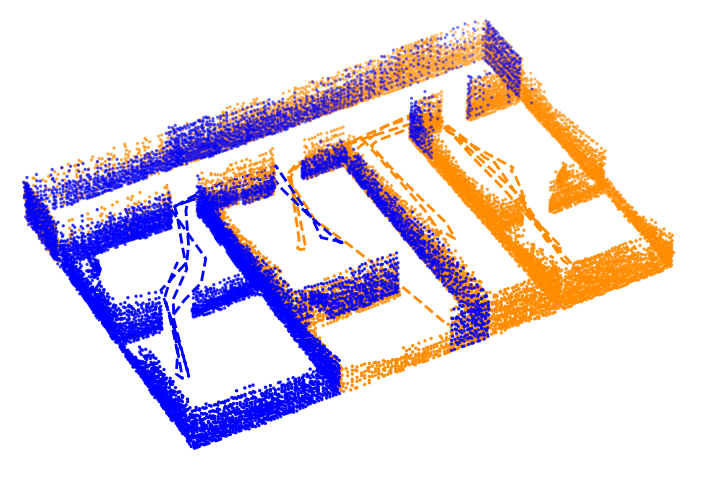}\\
        \caption{SE1}
        \label{fig:exp:sim:SE1}
     \end{subfigure}
        \caption{Simulated experiments, showing the trajectory and the point cloud map generated for each robot.}
        \label{fig:exp:sim:pcs}
\end{figure}

\textbf{Real-world Data.}
The datasets utilized in our experiments consist of a collection of in-house datasets acquired from a construction site: C1F1, C1F2, C3F1, C3F2, and C2F2. In addition, an experiment with 3 robots was performed on the C2F2 stage enabled by the size of the scenario. 
\tb{For these experiments we relied on the robot encoders to generate the odometry data.}
Unfortunately, our real-world datasets do not include IMU data and consequently, algorithms that rely heavily on IMU, such as DCL-SLAM and Disco-SLAM, could not be evaluated.

\begin{figure}[!hb]
     \centering
\vspace{-0.2cm}
     \begin{subfigure}[b]{0.16\textwidth}
        \centering
        \includegraphics[width=0.98\textwidth]{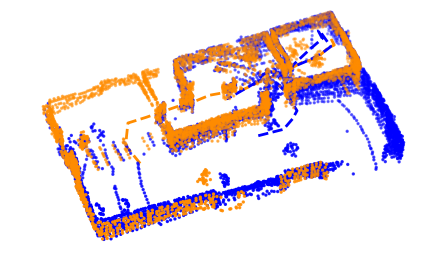}\\
        \caption{C1F1}
        \label{fig:exp:real:C1F1}
     \end{subfigure}
     \begin{subfigure}[b]{0.16\textwidth}
        \centering
        \includegraphics[width=0.98\textwidth]{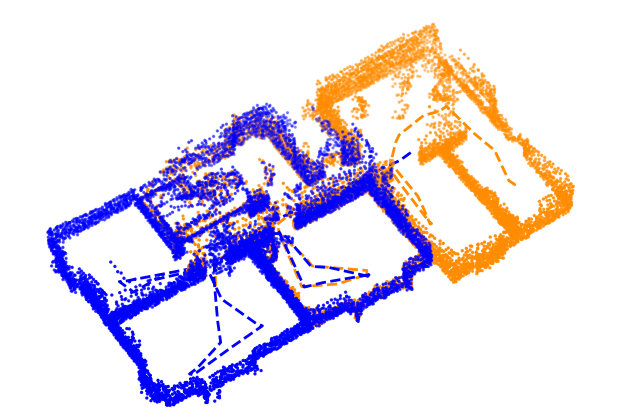}\\
        \caption{C1F2}
        \label{fig:exp:real:C1F2}
     \end{subfigure}
     \begin{subfigure}[b]{0.15\textwidth}
        \centering
        \includegraphics[width=0.98\textwidth]{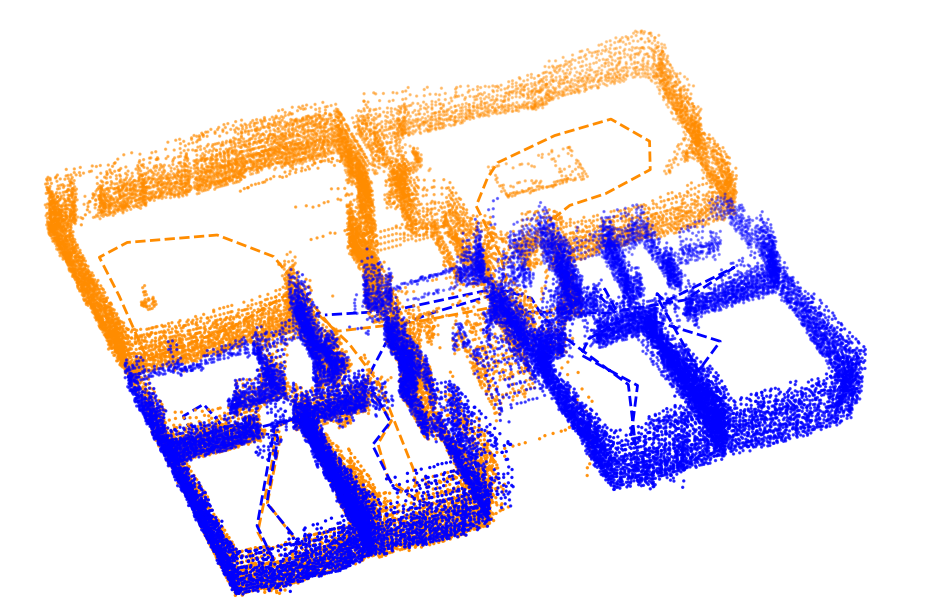}\\
        \caption{C3F1}
        \label{fig:exp:real:C3F1}
     \end{subfigure}
     \begin{subfigure}[b]{0.17\textwidth}
        \centering
        \includegraphics[width=0.9\textwidth]{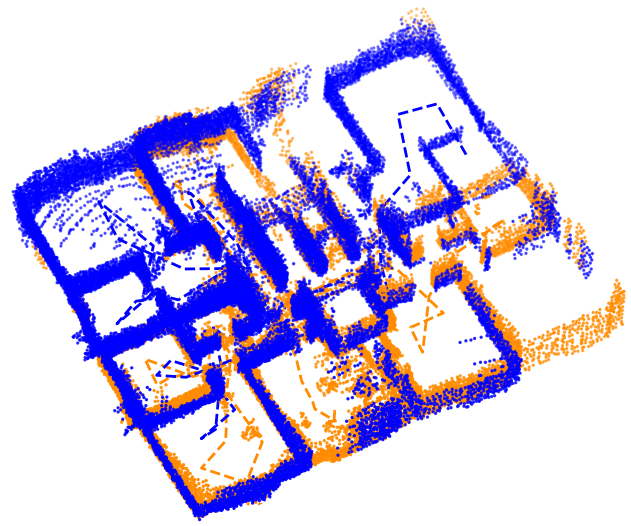}\\
        \caption{C3F2}
        \label{fig:exp:real:C3F2}
     \end{subfigure}
     \begin{subfigure}[b]{0.30\textwidth}
        \centering
        \includegraphics[width=0.98\textwidth]{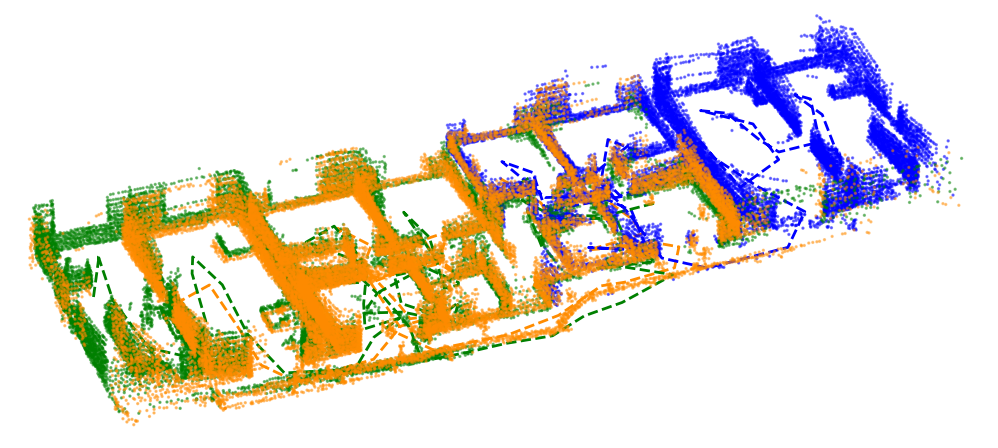}\\
        \caption{C2F2}
        \label{fig:exp:real:C2F2}
     \end{subfigure}
        \caption{Real experiment scenarios, showing the trajectory and the point cloud map generated for each robot. In (e) the experiment was run with 3 robots.}
        \vspace{-0.3cm}
        \label{fig:exp:real:pcs}
\end{figure}
\textbf{Ablation Study.}
In this study, we performed a thorough analysis of the contribution of the use of hierarchical graph factors (Section~\ref{sec:collaborative_s_graphs}) and the influence of the use of Room Centric point cloud (Section~\ref{sec:room_descriptors}) in the precision of the graph alignment, along with the bandwidth exchanged. 
The ablated stages are described below.

\textit{\textbf{Rooms:}} The matched room entities between robots are connected using room-to-room edges without fine alignment.

\textit{\textbf{Rooms with Fine Alignment:}} Room-to-room matching is improved with fine alignment between the room point clouds.

\textit{\textbf{Rooms and Walls:}} Walls are added to the room-to-room connection using wall-to-wall edges without fine alignment.

\textit{\textbf{Rooms and Walls with Fine Alignment (Full algorithm):}} Room-to-room matching with wall-to-wall edges and fine alignment.

When \textit{Fine Alignment} (Subsection \ref{section:loop_closure}) is not performed in loop closure, no point cloud information of the room is exchanged between robots.



\subsection{Experimental Setup}
The robot used for generating the datasets is a Boston Dynamics Spot with a Velodyne VLP-16 onboard. The experiments were carried out using a laptop computer with an Intel i7-10870H (8 cores, 2.2 GHz) with 32 GB of RAM memory. \textit{Multi S-Graphs} is developed under \textit{S-Graphs+}\cite{Bavle2022} using ROS 2 for implementing the algorithms.
\tb{The room matching parameters have been tuned to give a robust performance in the experimental environments, obtaining values of $SC_{th}=0.35$ and $ICP_{th}=0.07$.}

\subsection{Results and Discussions}


\textbf{Simulated Data.}

Fig. \ref{fig:exp:sim:pcs} shows the qualitative results of \textit{Multi S-Graphs} obtained during the simulated experiments. 
The outcomes presented in Table \ref{tab:exp:sim:ate} show that \textit{Multi S-Graph} algorithm exhibits a lower Average Trajectory Error (ATE) for the aggregated robot trajectory compared with the selected state-of-the-art (SOTA) CSLAM algorithms. Our approach outperformed all other decentralized algorithms, reducing the accuracy error by more than $90\%$ ($93.7\%$, $94.5\%$, and $98.4\%$ for DCL-SLAM, Disco-SLAM, and Swarm-SLAM respectively). It also achieved better accuracy than LAMP 2.0, with $72.9\%$ less error, which requires precise initial poses of the robots. The lack of features in the simulation environment causes measurement errors to adversely affect optimization. The extraction of semantic information specific to this type of environment by our algorithm overcomes this source of error.

In addition, our approach reduces the exchange of information between the agents by $94.50\%$, $97.20\%$, and $97.64\%$ for Swarm-SLAM, DCL-SLAM, and Disco-SLAM, respectively, and $81.3\%$ for LAMP 2.0 (see Table \ref{tab:exp:sim:mb}). This can be attributed to the effective use of semantic information to reduce the amount of data exchanged.

\begin{table}[!htb]
    \centering
    \setlength{\tabcolsep}{3pt}
    \begin{tabular}{l | rrr | r}
         \toprule
          \textbf{ATE [cm]} & \multicolumn{4}{c}{\textbf{2 Robots}} \\
          \midrule
         {\textbf{Algorithm}} & C1F0s & C1F2s & SE1 & \multicolumn{1}{c}{Avg}\\
         \toprule
         LAMP 2.0   & 9.66 & 8.10 & 6.27  & 8.01 $\pm$ 1.69\\
         DCL-SLAM   & 31.15 & 33.81 & 38.43 & 34.46 $\pm$ 3.68\\
         Disco-SLAM & 35.81 & 38.45 & 45.03 & 39.76 $\pm$ 4.75 \\
         Swarm-SLAM & 119.95 & 78.08  & 213.09 & 137.04 $\pm$ 69.11  \\
         \textit{Multi S-Graphs (ours)} & \textbf{3.64} & \textbf{1.39} & \textbf{1.50} &  \textbf{2.17} $\pm$ \textbf{1.26}\\
        \bottomrule
    \end{tabular}
    \caption{Absolute Trajectory Error (ATE) [cm] for our simulation experiments. The best results are boldfaced.}    
    \label{tab:exp:sim:ate}
\end{table}


\begin{table*}[ht]
    \centering
    \vspace{0.2cm}
    \begin{tabular}{l | r  r  r  r  r | r | r}
        \toprule
        \textbf{Point Cloud RMSE [cm]} & \multicolumn{6}{c|}{\textbf{2 Robots}} & \multicolumn{1}{c}{\textbf{3 Robots}} \\
        \midrule
        \textbf{Algorithm} &  C1F1 & C1F2 & C2F2 & C3F1 & C3F2 & \multicolumn{1}{c|}{Avg} & C2F2  \\
        \midrule
        LAMP 2.0  & 61.49 & 23.61 & 31.89 & 30.46 & 40.05 & 37.50 $\pm$ 14.63 & 35.34 \\
        Swarm-SLAM & 75.45 & 94.60 &147.22 & 40.36 & 36.42 & 78.81 $\pm$ 45.32 & 158.60 \\
        \textit{Multi S-Graphs (ours)} & \textbf{27.03}   &\textbf{15.12}  &\textbf{17.73}  & \textbf{21.04} & \textbf{21.58} & \textbf{20.18} $\pm$ \textbf{4.49} & \textbf{20.86}\\
        \bottomrule
    \end{tabular}
    \caption{Point cloud RMSE [cm] for our in-house real experiments. The best results are boldfaced.}
    \vspace{-0.2cm}
    \label{tab:exp:real:rmse}
\end{table*}

\begin{table}[!htb]
    \centering
    \begin{tabular}{l|rrrc}
        \toprule
         \textbf{Data Exchange} [\textbf{MB}] & \multicolumn{3}{c}{\textbf{2 Robots}} \\ 
         \midrule
          {\textbf{Algorithm}} & C1F0s & C1F2s & SE1\\
         \toprule
         LAMP 2.0  & 3.23& 3.69& 7.68\\
         DCL-SLAM & 25.20&24.89&41.54\\
         Disco-SLAM & 30.28 & 32.36 & 45.69\\
         Swarm-SLAM & 11.31&14.40&20.80\\
         \textit{Multi S-Graphs (ours)} & \textbf{0.53} & \textbf{0.95} & \textbf{1.08}\\
        \bottomrule
    \end{tabular}
    \caption{Data exchanged [MB] between agents in simulation experiments. The best results are boldfaced.}    
    \label{tab:exp:sim:mb}
    \vspace{-0.4cm}
\end{table}
\textbf{Real-world Data.}
Table \ref{tab:exp:real:rmse} presents the outcomes of the real-world experiments. Results from real datasets resemble the previous simulation results, with our algorithm showing $78.81\%$ improved performance over Swarm-SLAM in map matching accuracy, and $45.3\%$ over LAMP 2.0. In this case, although the accuracy of the extracted semantic features slightly decreases due to other elements present in real environments, the use of semantic-relational information of room-walls and their point clouds allows our approach to obtain better results than the other algorithms, achieving average reconstruction error of 20.5 cm that slightly vary in the different scenarios. 

\begin{table}[!htb]
    \setlength{\tabcolsep}{4pt}
    \centering
    \vspace{0.1cm}
    \begin{tabular}{l | rrrrr | r}
         \toprule 
          \textbf{Data Exchange} [\textbf{MB}] & \multicolumn{5}{c|}{\textbf{2 Robots}} & \multicolumn{1}{c}{\textbf{3 Robots}} \\ 
          \midrule
         \textbf{Algorithm} &  C1F1 & C1F2 & C2F2 & C3F1 & C3F2 & C2F2\\
         \toprule
         LAMP 2.0  &  1.89& 3.72 & 19.03& 8.18& 8.59 & 20.48 \\
         Swarm-SLAM &  4.31& 8.80 & 44.84& 17.53& 17.65 & 48.79 \\
        \textit{Multi S-Graphs (ours)} & \textbf{0.26} & \textbf{0.82} & \textbf{3.50}    & \textbf{1.12}   &  \textbf{0.66}& \textbf{5.73}\\
        \bottomrule
    \end{tabular}
    \caption{Data exchanged [MB] between agents in real experiments. The best results are boldfaced.}
    \label{tab:exp:real:mb}
    \vspace{-0.2cm}
\end{table}
Validation in real experiments concludes what has been demonstrated in simulations, with our approach achieving a high level of precision in reconstructing the building map (see Fig. \ref{fig:exp:real:pcs}). 
Table \ref{tab:exp:real:mb} shows the amount of data exchanged between agents. In this case, our approach reduces by $84.89\%$ and $93.34\%$ the amount of data exchanged between robots compared to LAMP 2.0 and Swarm-SLAM, respectively. 

\textbf{Ablation Study.}
The results of this study are collected in Table \ref{tab:exp:ablation}. 
We can see that the utilization of hierarchical optimization, which establishes relationships between the walls of two rooms among agents (\textit{Rooms and Planes}), means a significant improvement of accuracy, reducing the ATE by $25.6\%$ with an increase in the amount of data by $77\%$. On the other hand, the use of Fine alignment improves the performance by $73.6\%$ and $25.6\%$ over the \textit{Rooms} and \textit{Rooms and Walls} configurations, respectively, with a 40 times higher data exchange requirement. 
In general, fine alignment techniques perform better compared to their non-fine alignment counterparts, primarily due to the limited angular resolution of the Scan Context, which is insufficient for precise alignment (see Section~\ref{sec:room_descriptors}). Furthermore, the use of wall-to-wall factors contributes to alignment improvement, consequently significantly enhancing overall performance. Consequently, the best performance is obtained by combining the semantic information extracted from the walls with fine alignment using the Room Centric point clouds, \textit{Rooms and Walls with Fine alignment}, improving the performance by up to $74.7\%$. However, due to the larger amount of data exchanged required, when communication channels are limited, \textit{Rooms and Walls}, which only use semantic information and avoid sending point cloud messages, may be an interesting solution without a significant decrease in accuracy.


\begin{table}[!h]
    \vspace{-0.1cm}
    \setlength{\tabcolsep}{2.8pt}
    \centering
    \begin{tabular}{l | r |  r | r | r | r | r}
         \toprule
          & \multicolumn{2}{c|}{\textbf{C1F0s}} & \multicolumn{2}{c|}{\textbf{C1F2s}} &  \multicolumn{2}{c}{\textbf{SE1}} \\ 
          \midrule
         \textbf{Algorithm} & \textbf{ATE}  &  \textbf{\makecell{Data \\ Shared}}  & \textbf{ATE}  & \textbf{\makecell{Data \\ Shared}}   & \textbf{ATE}  & \textbf{\makecell{Data \\ Shared}}  \\
         \midrule
         \textit{Rooms }  & 9.78 & \textbf{12.88} &1.99&\textbf{18.90}&8.06&\textbf{26.46}\\
         \textit{Rooms with FA}  & 3.78 &  524.32 &1.42&946.91&1.62&1069.03\\
         \textit{Rooms $\&$ Walls}  & 5.33 & 22.81 &1.72&27.38&1.73&37.85\\
         \textit{Rooms $\&$ Walls with FA}  & \textbf{3.64} & 537.53 & \textbf{1.39}  &  955.39 & \textbf{1.50} & 1080.41 \\

        \bottomrule
    \end{tabular}
\caption{Ablation study of the ATE  [cm] and data shared [MB] for simulated datasets with 2 robots, FA means Fine Alignment. The best results are boldfaced.}
\label{tab:exp:ablation}
\vspace{-0.2cm}
\end{table}

\begin{figure}[!htb]
    \vspace{-0.3cm}
    \centering
    \includegraphics[width=0.4\textwidth]{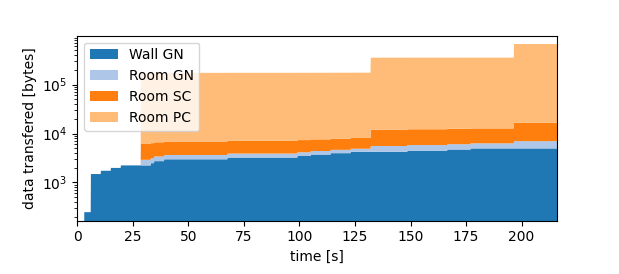}
    \caption{Accumulated data shared between agents, consisting of Walls Nodes (Wall GN) and Room Nodes (Room GN) for the Semantic Graph, and the Room Scan Contexts (Room SC) and Room Centric point clouds (Room PC). 
    }
    \vspace{-0.3cm}
    \label{fig:bandwith}
\end{figure}

In Fig. \ref{fig:bandwith} we can see on a logarithmic scale how the graph information (composed mainly of walls and room graph nodes) is negligible compared to the room point clouds. This observation highlights that point clouds lead to a substantial increase in transmitted data. Although the growth in the graph data during map exploration does not result in a significant increase in data volume, the cost associated with point clouds significantly escalates.

\section{Conclusions and Future Work}
In this work a distributed CSLAM system is presented, leveraging the semantic-relational features extracted by the S-Graphs \cite{bavle2022a}, \cite{Bavle2022}, in order to filter and reduce the amount of data that has to be transmitted between robots while outperforming state-of-the-art mapping and positioning accuracy in structured environments. 
\tb{We proposed a novel Room Descriptor that fuses the room structural elements with the fine-grained point cloud information, reducing the number of candidates for loop closure, without requiring the identification of objects present in the scene or exchanging keyframes between agents.}

We tested our algorithm for a map generation task in multiple scenarios, simulated and real, outperforming other state-of-the-art lidar-based CSLAM algorithms in terms of precision and bandwidth, reducing the trajectory and map reconstruction error by more than $90\%$ while saving $95\%$ of the data exchanged between robots.

The main drawback is the dependence on semantic-relational structures, defined in our case as four-walled rooms in building-like environments, which are not always easy to extract. \tb{The room matching relies on these room structures and the detection of different static features in them, so having a non-feature environment or changing environment can cause the room matching not to work properly.}

For future work, a more generic way \tb{of detecting different types of rooms and finding alignment between Situational Graphs should be found to increase the variety of scenarios in which \textit{Multi S-Graphs} can be used.}
\tb{Also, a validation of the system running onboard the robots must be performed to test the real-time capabilities of the system.}


\vspace{-0.1cm}

\bibliographystyle{IEEEtran}
\bibliography{bibliography}

\end{document}